\documentclass{article}
\usepackage{ijcai20}

\usepackage{times}
\usepackage{soul}
\usepackage{url}
\usepackage[hidelinks,colorlinks,linkcolor=red,citecolor=MyDarkBlue]{hyperref}
\usepackage[utf8]{inputenc}
\usepackage[small]{caption}
\usepackage{graphicx}
\usepackage{amsmath}
\usepackage{amsthm}
\usepackage{booktabs}
\usepackage{algorithm}
\usepackage{algorithmic}
\usepackage{courier}
\usepackage{color,xcolor}
\usepackage{epsfig}
\usepackage{graphicx}

\usepackage{adjustbox}
\usepackage{array}
\usepackage{booktabs}
\usepackage{colortbl}
\usepackage{float,wrapfig}
\usepackage{hhline}
\usepackage{multirow}
\usepackage{caption}
\usepackage{subfigure}

\usepackage{amsmath,amsfonts,amsthm,amssymb}
\usepackage{bm}
\usepackage{nicefrac}
\usepackage{microtype}

\usepackage{changepage}
\usepackage{extramarks}
\usepackage{fancyhdr}
\usepackage{lastpage}
\usepackage{setspace}
\usepackage{soul}
\usepackage{xspace}

\usepackage{enumerate}
\usepackage[draft]{todonotes} 

\usepackage{titlesec}

\usepackage{comment}
\usepackage{tikz}
\usetikzlibrary{bayesnet}

\newcolumntype{L}[1]{>{\raggedright\let\newline\\\arraybackslash\hspace{0pt}}m{#1}}
\newcolumntype{C}[1]{>{\centering\let\newline\\\arraybackslash\hspace{0pt}}m{#1}}
\newcolumntype{R}[1]{>{\raggedleft\let\newline\\\arraybackslash\hspace{0pt}}m{#1}}

\newcommand{\ignore}[1]{}

\makeatletter
\DeclareRobustCommand\onedot{\futurelet\@let@token\@onedot}
\def\@onedot{\ifx\@let@token.\else.\null\fi\xspace}

\def\ie{i.e\onedot}

\def\wrt{w.r.t\onedot} 

\makeatother

\definecolor{MyDarkBlue}{rgb}{0,0.08,0.45}
\definecolor{MyDarkGreen}{rgb}{0.02,0.6,0.02}
\definecolor{MyDarkRed}{rgb}{0.8,0.02,0.02}
\definecolor{MyDarkOrange}{rgb}{0.40,0.2,0.02}
\definecolor{MyPurple}{RGB}{111,0,255}
\definecolor{MyRed}{rgb}{1.0,0.0,0.0}
\definecolor{MyGold}{rgb}{0.75,0.6,0.12}
\definecolor{MyDarkgray}{rgb}{0.66, 0.66, 0.66}

\newcommand{\citein}[1]{\citeauthor{#1} \shortcite{#1}}

\title{DualSMC: Tunneling Differentiable Filtering and Planning under \\Continuous POMDPs}

\author{
Yunbo Wang$^1$\thanks{Equal contribution}
\and
Bo Liu$^{2*}$\and
Jiajun Wu$^3$\and
Yuke Zhu$^2$\and
Simon S. Du$^4$\and\\
Li Fei-Fei$^3$\And
Joshua B. Tenenbaum$^5$
\affiliations
$^1$Tsinghua University\\
$^2$The University of Texas at Austin\\
$^3$Stanford University\\
$^4$Institute for Advanced Study\\
$^5$Massachusetts Institute of Technology
\emails
yunbo.thu@gmail.com, bliu@cs.utexas.edu
}

\begin{document}

\maketitle

\begin{abstract}
A major difficulty of solving continuous POMDPs is to infer the multi-modal distribution of the unobserved true states and to make the planning algorithm dependent on the perceived uncertainty. We cast POMDP filtering and planning problems as two closely related Sequential Monte Carlo (SMC) processes, one over the real states and the other over the future optimal trajectories, and combine the merits of these two parts in a new model named the DualSMC network. In particular, we first introduce an \textit{adversarial particle filter} that leverages the adversarial relationship between its internal components. Based on the filtering results, we then propose a planning algorithm that extends the previous SMC planning approach \cite{piche2018probabilistic} to continuous POMDPs with an uncertainty-dependent policy. Crucially, not only can DualSMC handle complex observations such as image input but also it remains highly interpretable. It is shown to be effective in three continuous POMDP domains: the floor positioning domain, the 3D light-dark navigation domain, and a modified Reacher domain\footnote[2]{Code available
at \url{https://github.com/Cranial-XIX/DualSMC}}.
\end{abstract}

\section{Introduction}
\label{intro}
Partially Observable Markov Decision Processes (POMDPs) formulate reinforcement learning problems where the agent's instant observation is insufficient for optimal decision making~\cite{kaelbling1998planning}. Decision making with partial observations requires taking the history into account, which brings a high computation cost. It is a known result that finding the optimal policy in finite-horizon POMDPs is PSPACE-complete~\cite{papadimitriou1987complexity}, which makes POMDPs difficult to solve in moderately large discrete spaces, let alone \emph{continuous domains}.

Approximate solutions to POMDPs based on deep reinforcement learning can directly encode the history of past observations with deep models like RNNs~\cite{hausknecht2015deep,karkus2017qmdp,zhu2018improving,igl2018deep,hafner2018learning}. 
Learning is done in an end-to-end fashion and the resulting models can handle complex observations including visual inputs. 
However, since conventional POMDP problems usually present an \emph{explicit state formulation}, executing the planning algorithm in a latent space makes it difficult to adopt any useful prior knowledge. Besides, whenever these models fail to perform well, it is difficult to analyze which part causes the failure as they are less interpretable.

In this work, we present a simple but effective model named Dual Sequential Monte Carlo (DualSMC). It preserves high interpretability since the state belief is represented by particles in real state spaces. It is also flexible for solving continuous POMDPs with complex observations and unknown dynamics. The idea of DualSMC is inspired by the recent successes on differentiable particle filters~\cite{jonschkowski2018differentiable,karkus2018particle} and the \textit{control as inference} framework~\cite{kappen2012optimal,levine2018reinforcement,piche2018probabilistic}. 
In particular, DualSMC solves continuous POMDPs by connecting a SMC filter for state estimation with a SMC planner that samples in the optimal future trajectory space\footnote{To distinguish the two SMCs, we call the first state estimation SMC the \textit{filter} and its particles the \textit{state particles}, and the second planning SMC the \textit{planner} and its particles the \textit{trajectory particles}.}.

Since the performance of the planner significantly depends on that of the filter, we introduce a novel adversarial training method to enhance the filter. Moreover, to connect the two parts and reduce the computational burden, we feed the top candidates of the state particles into the planner as the initial belief for uncertainty-aware action selection. The planner also takes as input the mean of the top state particles. 
To further improve robustness, we perform the \textit{model predictive control} where only the first action of the plan is selected and we re-plan at each step.
Notably, the learned dynamics is efficiently \emph{shared} between filtering and model-based planning.

Our contributions to continuous POMDPs with DualSMC can be summarized as follows:
\begin{itemize}
    \item It proposes a new \emph{differentiable particle filter} (DPF) that leverages the adversarial relationship between the internals of the original DPF \cite{jonschkowski2018differentiable}.
    \vspace{-2pt}
    \item It introduces a new POMDP planning algorithm in forms of neural networks that extends the original sequential Monte Carlo planning \cite{piche2018probabilistic} from fully observable scenarios to partially observable ones. The algorithm ties the transition model between the filter and the planner, bridges them via particle-based belief states, and learns the uncertainty-aware policy from the belief.
    \vspace{-2pt}
    \item It provides new benchmarks for continuous POMDPs: the \emph{floor positioning} for explanatory purposes, the 3D \emph{light-dark} navigation with rich visual inputs, and a control task in the Mujoco environment \cite{Todorov2012MuJoCo}. DualSMC achieves the best results consistently.
\end{itemize}

\section{Problem Setup}\label{background}

A continuous POMDP can be usually specified as a 7-tuple $(\mathcal{S}, \mathcal{A}, \mathcal{T}, \mathcal{R}, \Omega, \mathcal{Z}, \gamma)$, where $\mathcal{S}$, $\mathcal{A}$ and $\Omega$ are continuous state, action and observation spaces. We denote $s_{t} \in \mathcal{S}$ as the underlying state at time $t$. When the agent takes an action $a_t \in \mathcal{A}$ according to a policy $\pi(a_t | o_{\leq t}, a_{< t})$, the state changes to $s_{t+1}$ with probability $\mathcal{T}(s_{t+1} | s_{t}, a_t)$. The agent will then receive a new observation $o_{t+1} \sim \mathcal{Z}(o_{t+1} | s_{t+1})$ and a reward $r_{t} \sim \mathcal{R}(s_t, a_t)$. 
Assuming the episodes are of fixed length $L$, the agent's objective is then to maximize the expected cumulative future reward
$G = \mathbb{E}_{\tau \sim \pi}\big[\sum_{t=1}^{L} \gamma^{t-1} r_t\big]$,
where $\tau = (s_1, a_1, \dots, a_L, s_{L+1})$ are trajectories induced by $\pi$, and $0 \leq \gamma < 1$ is the discount factor. 
Since observations generally do not reveal the full state of the environment, the classical methods often maintain a belief over possible states, $\text{bel}(s_t) \triangleq p(s_t | o_{\leq t}, a_{< t})$, and update the belief according to
\vspace{-2pt}
\begin{equation}
\small
\text{bel}(s_{t+1}) = \eta\int \text{bel}(s_t) \mathcal{Z}(o_{t+1} | s_{t+1}) \mathcal{T}(s_{t+1}|s_t, a_t) ds_t,
\end{equation}
where $\eta$ is a normalization factor. In this work, we make the true states available during training only as a supervised signal for the filter and keep them unobserved during testing. The key to solve continuous POMDPs is to perceive the state uncertainty and make decisions under such uncertainty.

\section{Related Work}
\label{related}

\paragraph{Planning under uncertainty.} Due to the high computation cost of POMDPs, many previous approaches used sampling-based techniques for either belief update or planning, or both. For instance, a variety of \textit{Monte Carlo tree search} methods have shown success in relatively large POMDPs by constructing a search tree of history based on rollout simulations~\cite{silver2010monte,somani2013despot,seiler2015online,sunberg2018online}. Later work further improved the efficiency by limiting the search space or reusing plans~\cite{somani2013despot,kurniawati2016online}. Although considerable progress has been made to enlarge the set of solvable POMDPs, it remains hard for pure sampling-based methods to deal with unknown dynamics and complex observations like visual inputs. Therefore, in this work, we provide one approach to combine the efficiency and interpretability of conventional sampling-based methods with the flexibility of deep learning networks for complex POMDP modeling.


\vspace{2pt}
\paragraph{Differentiable particle filter.} Ever since its invention~\cite{gordon1993novel}, the Particle Filter (PF), or Sequential Monte Carlo (SMC), has become a well-suited method for sequential estimation in complex non-linear scenarios. 
A large number of research has made progress on learning a flexible proposal distribution for SMC\footnote{The proposal distribution refers to the posterior distribution over the latent variables in an SMC. This should not be confused with the particle proposer model in this paper, which is a separate model that proposes possible state particles given observations.}. 
\citein{gu2015neural}
was one of the earliest that use a recurrent neural network to model the proposal distribution.
\citein{naesseth2018variational} and \citein{maddison2017filtering} further provided a variational framework that learns a good parameterized proposal distribution by optimizing the log estimator. 
Recently, \citein{karkus2018particle} and \citein{jonschkowski2018differentiable} introduced differentiable particle filters independently and applied them to localization problems with rich visual input. These approaches explicitly treat the proposal distribution as three interleaved neural modules: a proposer that generates plausible states, a transition model that simulates dynamics, and an observation model that does Bayesian belief update.
The filter in our model is based on~\cite{jonschkowski2018differentiable}, with an additional adversarial objective. 
\citein{kempinska2017adversarial} also proposed an adversarial training objective for SMC. But their objective is for learning the proposal distribution, while our method focuses more on mutually enhancing the proposer and observation model.

\paragraph{Planning as inference.} The framework of \textit{control as probabilistic inference} considers that selecting the optimal action is equivalent to finding the maximum posterior over actions conditioned on an optimal future~\cite{todorov2008general,toussaint2009robot,kappen2012optimal,levine2013variational}. We refer to \cite{levine2018reinforcement} as an explanatory review of these methods. 
\citein{piche2018probabilistic} extended this idea further to planning problems and propose the sequential Monte Carlo planning (SMCP), where the inference is done over optimal future trajectories. While most previous work focused on Markov Decision Processes (MDP) with full observation, we take one step further and apply the \textit{planning as inference} framework to POMDP problems.
On the other hand, compared with the existing Bayesian reinforcement learning literature on POMDPs \cite{ross2008bayes}, our work focuses more on deep reinforcement learning solutions to continuous POMDPs.

\section{Dual Sequential Monte Carlo Network}

In this section, we first introduce the adversarial particle filter that aims to mutually enhance the particle proposer model and the observation model. 
Then, we illustrate the design choice of our main algorithm and describe in detail how our method connects the two SMCs, during which we also introduce an alternative simpler formulation for SMCP~\cite{piche2018probabilistic}.

\begin{figure*}[t]
  \centering
  \includegraphics[width=0.9\textwidth]{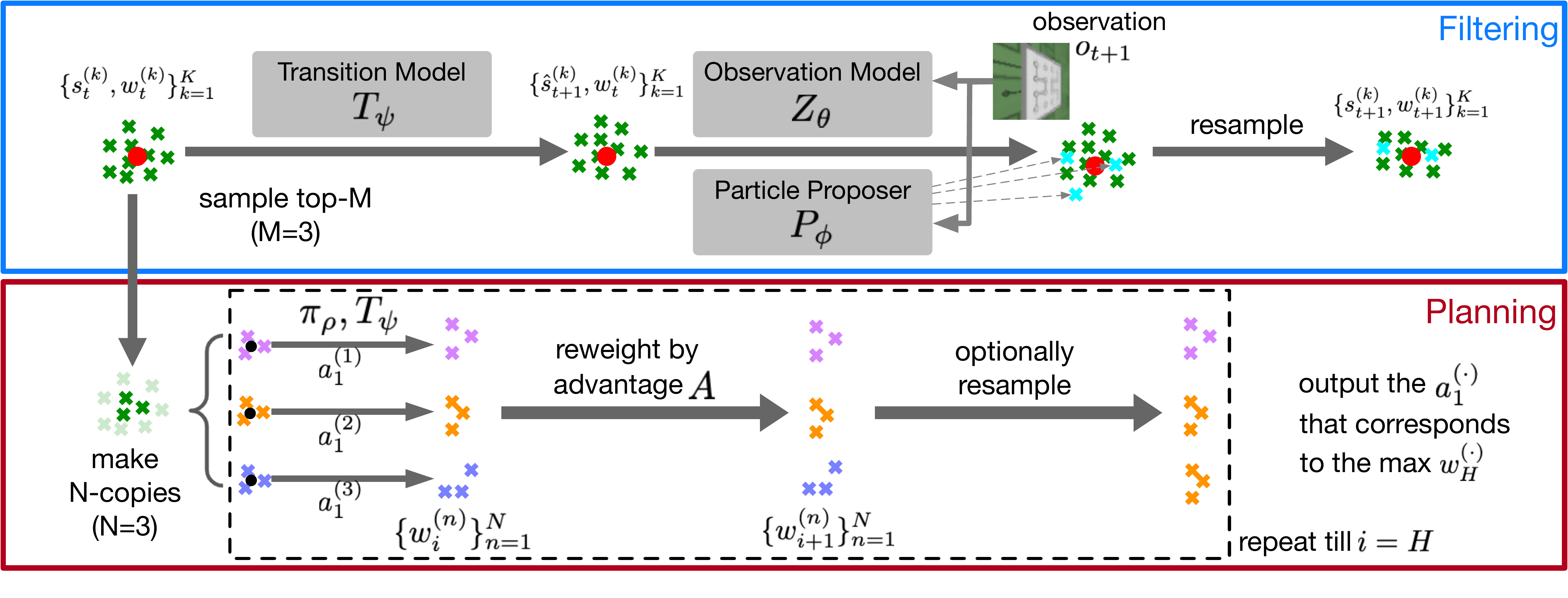}
  \vspace{-12pt}
  \caption{A schematic drawing of the modules in DualSMC. Here we choose $M=3$ and $N=3$ for illustration}
  \vspace{8pt}
  \label{fig:schematic}
\end{figure*}

\subsection{Adversarial Particle Filtering}
\label{sec:4.1}

A particle filter represents a \textit{belief distribution} $\text{bel}(s_t)$ of the true state $s_t$ with a set of weighted particles $\{ (s_t^{(k)}, w_t^{(k)}) \}_{k=1}^K$, where $\sum_{k=1}^{K} w_t^{(k)} = 1$. To perform Bayesian update when action is applied and a new observation comes in, it first transits all particles according to a transition model and then update corresponding weights according to an observation model:
\vspace{-2pt}
\begin{equation}
    s_{t+1}^{(k)} \sim \mathcal{T}(\cdot | s_t^{(k)}, a_t) \,\text{ and }\, w_{t+1}^{(k)} \propto \mathcal{Z}(o_{t+1} | s_{t+1}^{(k)})w_t^{(k)}.
\end{equation}

In practice, when the true dynamics $\mathcal{T}$ and $\mathcal{Z}$ are not known a priori, they can be approximated by the parameterized functions $T_\psi(\cdot)$ and $Z_\theta(\cdot)$. Similar to \cite{jonschkowski2018differentiable}, our differentiable particle filter contains three neural modules (Figure \ref{fig:schematic}): the proposer $P_\phi(o_t, \epsilon_P)$, the transition model $T_\psi(s_{t-1}^{(k)}, a_{t-1}, \epsilon_T)$, and the observation model $Z_\theta(o_t, s_t^{(k)})$, where $\phi, \psi, \theta$ are parameters. $\epsilon_P$ and $\epsilon_T$ are the Gaussian noises for stochastic models. 
To avoid the particle degeneracy problem \cite{doucet2009tutorial}, we perform Sequential Importance Resampling (SIR) together with the proposer model. Specifically, after the Bayesian update at time $t$, we sample $K^\prime$ old particles $\{s_\text{old}^{(k)}\}_{k=1}^{K^\prime}$ with replacement based on the updated weight and combine them with $(K-K^{\prime})$ newly proposed particles $\{s_\text{new}^{(k)}\}_{k=K^\prime+1}^{K}$, and assign uniform weights for all particles. Depending on the task, we keep $K^{\prime}$ constant or make $(K-K^{\prime})$ follow an exponential decay.

The major difference between our filtering approach and \cite{jonschkowski2018differentiable} comes in by noticing that $P_\phi$ and $Z_\theta$ are naturally opposite to yet dependent on each other. Following this intuition, instead of regressing the output of the proposer to the true state, we propose the adversarial proposing strategy. In particular, we train $Z_\theta$ to differentiate the true state from all particle states and train $P_\phi$ to fool $Z_\theta$. Formally, denote $p_\text{real}(o_{\leq t}), p_\text{real}(s|o_{\leq t})$ as the real distributions over observations and the real posterior over $s$, $Z_\theta$ and $P_\phi$ play the following two-player minimax game with function $F(Z_\theta,P_\phi)$:
\vspace{-10pt}
\begin{equation}
\label{eq:adv_observation_loss}
\begin{split}
  \min_\phi \max_\theta & F(Z_\theta,P_\phi) = \mathbb{E}_{o_{1:t} \sim p_{\text{real}(o_{\leq t})}} \Big[\\ 
  & \mathbb{E}_{s \sim p_{\text{real}(s | o_{\leq t})}} \log Z_\theta(o_t, s) \ + \\
  &\mathbb{E}_{s' \sim s_\text{old}^{(k)}} \log(1-Z_\theta(o_t, s’)) \ + \\
  &\mathbb{E}_{\epsilon_P \sim \mathcal{N}(0,I)} \log(1-Z_{\theta}(o_t, P_{\phi}(o_t,\epsilon_P)))\Big].
\end{split}
\end{equation}

During training, instead of using trajectories sampled from a random or heuristic policy \cite{jonschkowski2018differentiable,karkus2018particle}, we train the filter in an on-policy way so that it can take advantage of the gradually more powerful planner.

\subsection{DualSMC Planning on Explicit Belief States}
\label{sec:4.2}

A straightforward solution to POMDP planning is to train the planning module separately from the filtering module. At inference time, plans are made \emph{independently based on each particle state}. We thus name this planning algorithm the Particle-Independent SMC Planning (PI-SMCP) and use it as a baseline method. 
More details on PI-SMCP can be found in Appendix \ref{appendix_pismcp}. 
Although PI-SMCP is unbiased, it does not perform well in practice because it cannot generate policies based on dynamically varying state uncertainties.

We thus propose the DualSMC algorithm to explicitly consider the belief distribution by planning directly on an approximated belief representation, \ie, a combination of the top candidates from the filter (for computation efficiency) as well as the weighted mean estimate. We show the modules in DualSMC and how they relate to each other in Figure \ref{fig:schematic}.

\begin{algorithm}[t]
\small
\caption{Overall DualSMC algorithm} \label{alg:dualsmc}
\begin{algorithmic}[1]
\STATE $\{s_1^{(k)} \sim \text{Priori}(s_1)\}_{k=1}^K$, $\{w_0^{(k)} = 1\}_{k=1}^K$
\FOR{$t = 1:L$}
    \STATE // At each filtering and control step
    \STATE $\{w_{t}^{(k)} \propto w_{t-1}^{(k)} \cdot Z_\theta(s_{t}^{(k)}, o_{t})\}_{k=1}^K$
    \STATE $\overline{\text{bel}}_t = \sum_k w_t^{(k)} s_t^{(k)}$
    \STATE $\{\tilde{s}_t^{(m)}, \tilde{w}_t^{(m)}\}_{m=1}^M = \text{Top-M}(\{s_t^{(k)}, w_t^{(k)}\}_{k=1}^K), \text{\wrt} \{w_t^{(k)}\}_k$
    \vspace{-8pt}
    \STATE $a_t = \textbf{\text{DualSMC-P}}(\overline{\text{bel}}_t, \{\tilde{s}_t^{(m)}, \tilde{w}_t^{(m)}\}_{m=1}^M; \pi_\rho, Q_\omega)$
    \STATE $o_{t+1}, r_t \sim p_\text{env}(a_t)$
    \IF{resample}
    \STATE $\{s_t^{(k)}\}_{k=1}^{K^\prime} \sim \text{Multinomial}(\{s_t^{(k)}\}_{k=1}^K), \text{\wrt} \{w_t^{(k)}\}_k$
    \STATE $\{s_t^{(k)} \sim P_\phi(o_t)\}_{k=K^\prime+1}^K, \{w_t^{(k)} = 1\}_{k=1}^K$
    \vspace{2pt}
    \ENDIF
    \vspace{2pt}
    \STATE $\{s_{t+1}^{(k)} \sim T_\psi(s_t^{(k)}, a_t)\}_{k=1}^K$
    \vspace{-1pt}
    \STATE Add ($s_t, s_{t+1}, a_t, r_t, o_{t}, \overline{\text{bel}}_t, \{\tilde{s}_t^{(m)}, \tilde{w}_t^{(m)}\}_{m=1}^M$) to a buffer
    \STATE Sample a batch from the buffer and update ($\rho, \omega, \theta, \psi, \phi$)
\ENDFOR
\end{algorithmic}
\end{algorithm}

\begin{algorithm}[h]
\small
\caption{DualSMC planner on estimated belief states} \label{alg:dualsmc-p}
\textbf{Input:} $\overline{\text{bel}}_t$, $\{\tilde{s}_t^{(m)}, \tilde{w}_t^{(m)}\}_{m=1}^M$ \\
\textbf{Output}: $a_t$
\begin{algorithmic}[1]
\STATE $\{\tilde{w}_t^{(m)}\}_{m=1}^M = \text{Normalize}(\{\tilde{w}_t^{(m)}\}_{m=1}^M)$
\STATE $\big\{\{\hat{s}_t^{(m)(n)} = \tilde{s}_t^{(m)}\}_{m=1}^M, \hat{w}_{t-1}^{(n)} = 1, \overline{\text{bel}}_t^{(n)} = \overline{\text{bel}}_t \big\}_{n=1}^N$
\vspace{2pt}
\FOR{$i = t:t+H$}
    \STATE // At each planning time step
    \STATE $\big\{a_i^{(n)} \sim \pi_\rho (\{\hat{s}_i^{(m)(n)}\}_{m=1}^M, \overline{\text{bel}}_i^{(n)})\big\}_{n=1}^N$
    \STATE $\big\{\hat{s}_{i+1}^{(m)(n)}, r_i^{(m)(n)} \sim T_\psi(\hat{s}_i^{(m)(n)}, a_i^{(n)})\big\}_{m=1,n=1}^{M,N}$
    \vspace{-2pt}
    \STATE $\big\{\overline{\text{bel}}_{i+1}^{(n)} = \sum_m \tilde{w}_t^{(m)} \hat{s}_{i+1}^{(m)(n)} \big\}_{n=1}^N$
    \STATE $\big\{\hat{w}_{i}^{(n)} \propto \hat{w}_{i-1}^{(n)} \cdot \exp\big(\sum_m \tilde{w}_t^{(m)} A^{(m)(n)}\big)\big\}_{n=1}^N$
    \STATE $\big\{x_i^{(n)} = (\{\hat{s}_{i+1}^{(m)(n)}, \hat{s}_i^{(m)(n)}\}_{m=1}^M, \overline{\text{bel}}_{i+1}^{(n)}, a_i^{(n)})\big\}_{n=1}^N$
    \vspace{2pt}
    \IF{resample}
    \STATE $\big\{x_{t:i}^{(n)}\}_{n=1}^N \sim \text{Multinomial}(\{x_{t:i}^{(n)}\}_{n=1}^N), \text{\wrt} \{\hat{w}_{i}^{(n)}\big\}_n$
    \STATE $\big\{\hat{w}_{i}^{(n)} = 1\big\}_{n=1}^N$
    \vspace{2pt}
    \ENDIF
\ENDFOR
\STATE $a_t =$ first action of $x_{t:t+H}^{(n)}$, where $n \sim \text{Uniform}(1, \dots, N)$
\end{algorithmic}
\end{algorithm}

The overall algorithmic framework of DualSMC is shown in Alg \ref{alg:dualsmc}. At time step $t$, when a new observation comes, we first use the observation model $Z_\theta$ to update the particle weights (line 4 in Alg \ref{alg:dualsmc}), and then perform the DualSMC planning algorithm in Alg \ref{alg:dualsmc-p}. 
We duplicate the top-$M$ particles (for computation efficiency) and the mean belief state $N$ times as the root states of $N$ planning trajectories (line 1-2 in Alg \ref{alg:dualsmc-p}).
Different from the previous SMCP~\cite{piche2018probabilistic} method under full observations, the policy network $\pi_\rho$ perceives the belief states and predicts an action based on the top-$M$ particle states as well as the mean belief state (line 5 in Alg \ref{alg:dualsmc-p}).
We then perform $N$ actions to $M \times N$ states and use $T_\psi$ to predict the next states and rewards (line 6 in Alg \ref{alg:dualsmc-p}).
Since future observations $o_{>t}$ are not available at current time step, inspired by QMDP~\cite{littman1995learning}, we assume the uncertainty disappears at the next step, and thus approximate $\overline{\text{bel}}_{i>t}^{(n)}$ using the top-$M$ transition states as well as a set of fixed filtering weights (line 7 in Alg \ref{alg:dualsmc-p}).
We update the planning weight of each planning trajectory by summarizing the advantages of each state using the initial $M$ belief weights (line 8 in Alg \ref{alg:dualsmc-p}). 
Here, we introduce an alternative advantage formulation that is equivalent to the one used in \cite{piche2018probabilistic}:
\begin{equation}
\label{eq:advantage}
\small
\begin{split}
    & \text{TD}^{(m)(n)}_{i-1} = Q_\omega(\hat{s}_i^{(m)(n)}, a_i^{(n)}) - Q_\omega(\hat{s}_{i-1}^{(m)(n)}, a_{i-1}^{(n)}) + r_{i-1}^{(m)(n)}, \\
    & A^{(m)(n)} = \text{TD}^{(m)(n)}_{i-1} - \log \pi_\rho (a_i^{(n)} | \{\hat{s}_i^{(m)(n)}\}_{m=1}^M, \overline{\text{bel}}_i^{(n)}).
\end{split}
\end{equation}
\vspace{-5pt}

At time $t$, $Q_\omega(\hat{s}_{i-1}^{(m)(n)}, a_{i-1}^{(n)})$ and $r_{i-1}^{(m)(n)}$ are set to $0$. 
It is a benefit, because according to Eq. (3.3) in the SMCP paper \cite{piche2018probabilistic}, the advantage in SMCP depends on both $Q$ and the log expectation of $V$, which can be difficult to estimate accurately, while our approach only requires $Q$, which is much simpler.
We leave the full derivation to Appendix \ref{appendix_dev}.

At the end of each planning time step, we may apply resampling when necessary over $N$ planning trajectories\footnote{The resampling steps in both filtering and planning algorithms may not be performed at every time step (more details in Table \ref{tab:training_parameters}).} (line 11-12 in Alg \ref{alg:dualsmc-p}.).
When the planning horizon is reached, where $i=t+H$, we sample one planning trajectory (line 15 in Alg \ref{alg:dualsmc-p}) and feed its first action to the environment.
We then go back to the filtering part and update the belief states by resampling, proposing, and predicting the next-step particle states (line 9-13 in Alg \ref{alg:dualsmc}). 
Lastly, we train all modules of DualSMC, including the policy network, the critic network, and the three modules of the adversarial particle filter (line 14-15 in Alg \ref{alg:dualsmc}).

\section{Experiment}
\label{experiment}

Our experiments are designed in the following way. First, we use a 2D Floor Positioning task to illustrate the effectiveness of both the adversarial proposing and that of the uncertainty-dependent planning. Next, we test the DualSMC network on a much harder navigation task with both high uncertainty and visual input. At least, we present a modified Reacher environment to further test DualSMC's performance beyond the navigation domain.
All models are trained with the Adam optimizer \cite{kingma2014adam}. The training hyper-parameters are shown in Table \ref{tab:training_parameters}. 
We tuned the number of filtering particles as it largely determines the quality of belief state estimation, which is the foundation of the subsequent DualSMC planning algorithm. After many trials, we finally set it to $100$ for a balance between planning results and efficiency.  All experimental results are averaged over $5$ runs of training. 
Network details can be found in Appendix \ref{appendix_model}.

\begin{table}[t]
    \centering
    \begin{tabular}{llll}
    \toprule
    Hyper-parameter & A & B & C \\
    \midrule
    Training episodes & $10{,}000$ & $2{,}000$ & $5{,}000$ \\
    Learning rate & $0.001$ & $0.0003$ & $0.0003$ \\
    Batch size & $64$ & $128$ & $256$ \\
    \midrule
    Filtering particles & $100$ & $100$ & $100$ \\
    - Resampling frequency & $8$ & $3$ & $2$ \\
    \midrule
    Planning trajectories & $30$ & $10$ & $10$ \\
    - Planning time horizon & $10$ & $10$ & $1$ \\
    - Resampling frequency & $3$ & $1$ & $1$ \\
    \bottomrule
    \end{tabular}
    \vspace{-5pt}
    \caption{Training hyper-parameters for the (A) floor positioning, (B) 3D dark-light, and (C) modified reacher domains}
    \label{tab:training_parameters}
    \vspace{10pt}
\end{table}

\begin{figure}[t]
  \centering
  \includegraphics[width=\columnwidth]{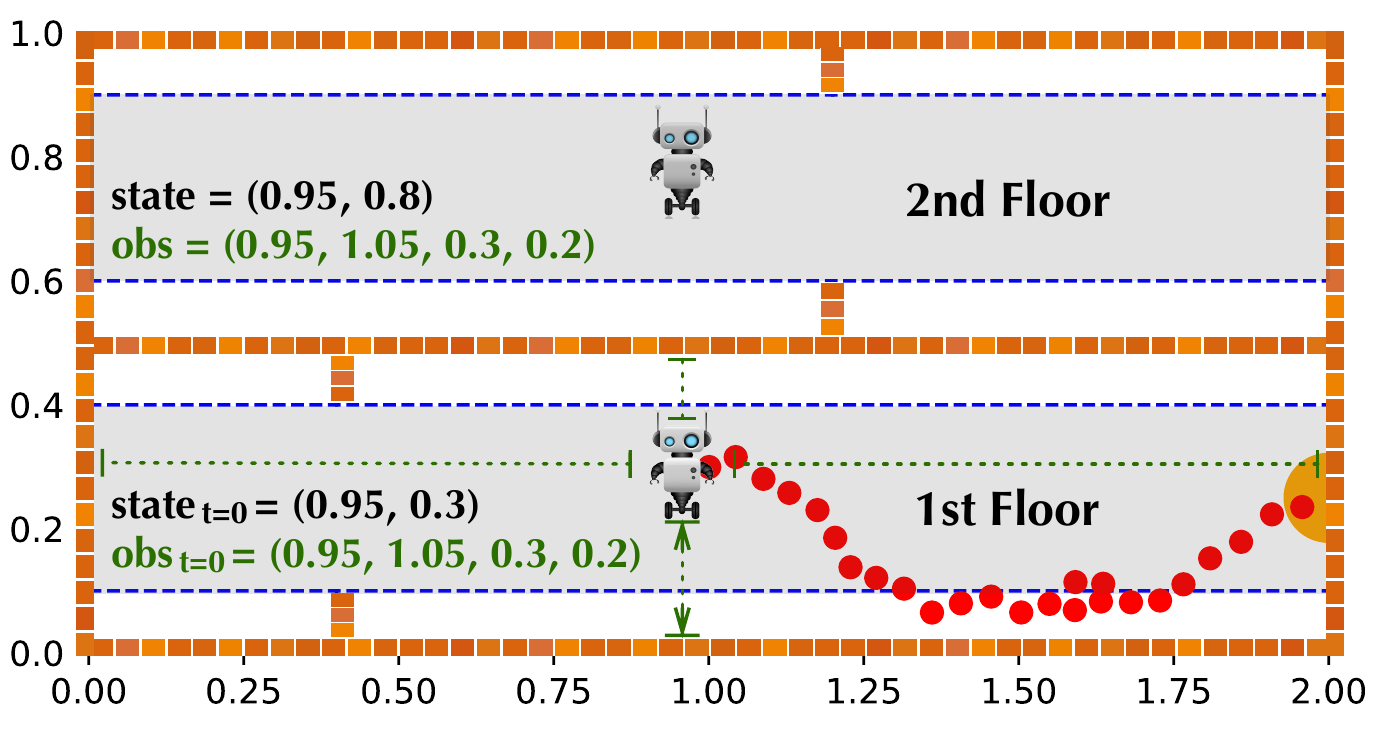}
  \vspace{-20pt}
  \caption{An illustration of the floor positioning domain}
  \label{fig:floor_position}
  \vspace{5pt}
\end{figure}

\begin{figure*}[t]
  \centering
  \includegraphics[width=\textwidth]{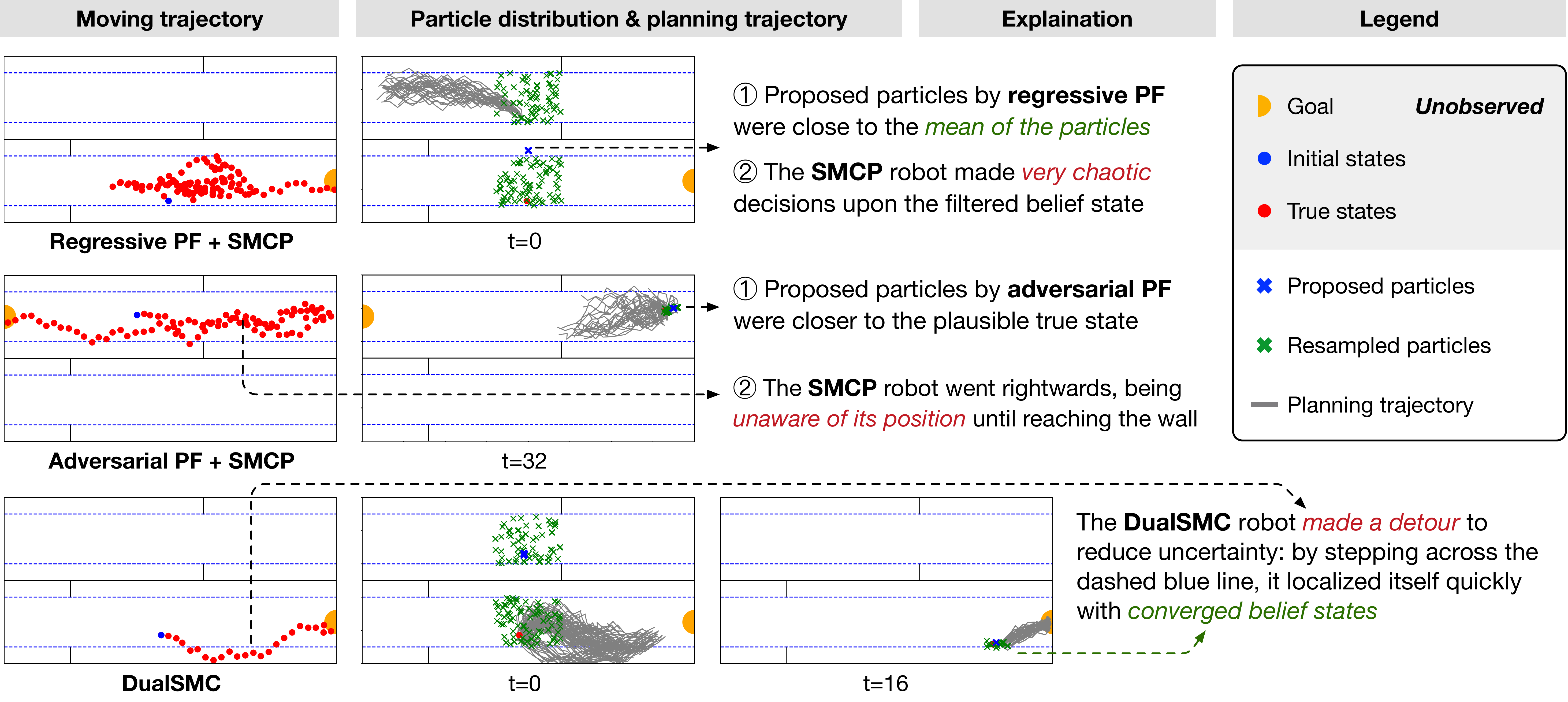}
  \vspace{-15pt}
  \caption{Qualitative results in the floor positioning domain, including the robot's actual moving trajectories and its planning trajectories
  }
  \label{fig:floor}
  \vspace{10pt}
\end{figure*}

\subsection{Floor Positioning}

Suppose there is a robot in a two-floor building as shown in Figure \ref{fig:floor_position}, who doesn't know which floor it resides on. It can only distinguish different floors by observation.
\begin{itemize}
    \item State: It is defined as the robot's position in world coordinates $(s_x,s_y)$, \ie, the axes in Figure \ref{fig:floor_position}.
    \item Action: It is defined as $a = (\Delta{s_x}, \Delta{s_y})$ with a maximum magnitude of $0.05$.
    \item Observation: It is defined as the robot's horizontal distances to the nearest left/right walls, and the vertical distances to ceiling/ground $o_t={(d_{x-}, d_{x+}, d_{y-}, d_{y+})}_t$. In the case of Figure \ref{fig:floor_position}, it starts with an observation of $(0.95,1.05,0.3,0.2)$, whatever floor it is on. 
    \item Goal: The robot starts from a random position and is headed to different regions according to different floors. If it is on the first floor, the target area is around $(2, 0.25)$ orange semicircle area); If the robot is on the second floor, the target area is around $(0, 0.75)$. Only at training time, a reward of $100$ is given at the end of each episode if the robot reaches the correct target area.
\end{itemize}

Starting from a gray area, the robot is very uncertain about its y-axis position. In the case of Figure \ref{fig:floor_position}, the estimated state can be $(0.95,0.3)$ or $(0.95,0.8)$. Only when the robot goes across a dashed blue line, from the gray area to the bright one, does it become certain about its y-axis position.

\begin{table}[t]
  \centering
  \resizebox{\columnwidth}{!}{
  \begin{tabular}{lcr}
    \toprule
    Method & Success & \# Steps \\
    \midrule
    DVRL \cite{igl2018deep} & 38.3\% & 162.0 \\
    \midrule
    LSTM filter + SMCP \cite{piche2018probabilistic} & 23.5\% & 149.1 \\
    Regressive PF ($\ell_2$, top-1) + SMCP & 25.0\% & 107.9 \\
    Regressive PF (density, top-3) + PI-SMCP & 25.0\% & 107.9 \\
    \midrule
    Adversarial PF (top-1) + SMCP & 95.0\% & 73.3 \\
    Adversarial PF (top-3) + PI-SMCP & 82.7\% & 86.9 \\
    \midrule
    DualSMC with regressive PF ($\ell_2$) & 45.1\% & 114.9\\
    DualSMC with regressive PF (density) & 58.3\% & 107.0 \\
    DualSMC w/o proposer & 78.6\% & 62.1 \\
    DualSMC with adversarial PF & \bf 99.4\% & \bf 26.9 \\
    \bottomrule
  \end{tabular}
  }
  \vspace{-5pt}
  \caption{The success rate and the average number of steps of $1{,}000$ tests in the floor positioning domain (PF is short for particle filter)}
  \vspace{10pt}
  \label{tab:toy}
\end{table}

\begin{figure}[t]
\vspace{-12pt}
  \centering
  \subfigure[Different POMDP planners]{
  \includegraphics[width=0.472\columnwidth]{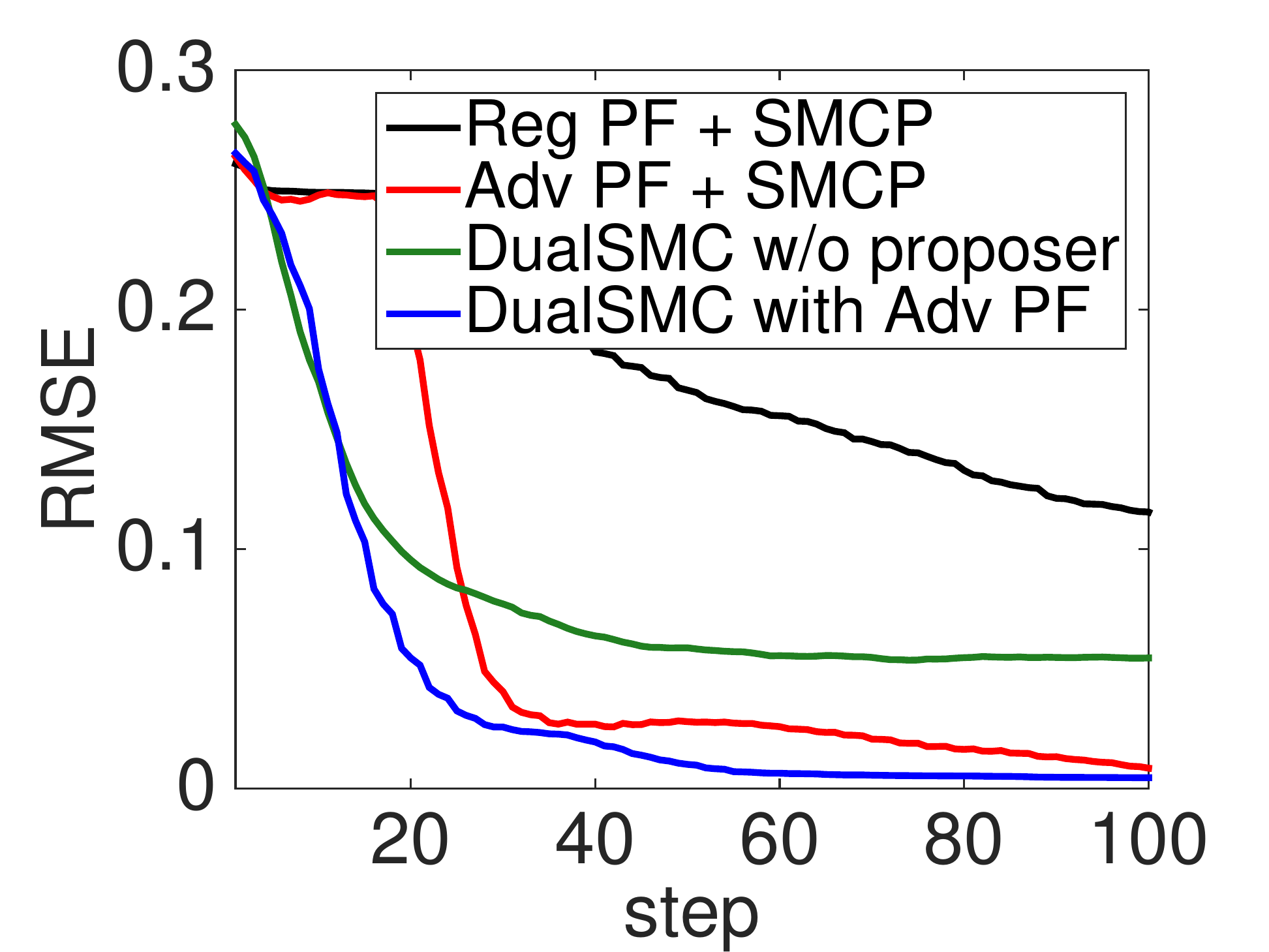}
  \label{fig:rmse_model_compare}
  }
  \subfigure[Different particle filters]{
  \includegraphics[width=0.472\columnwidth]{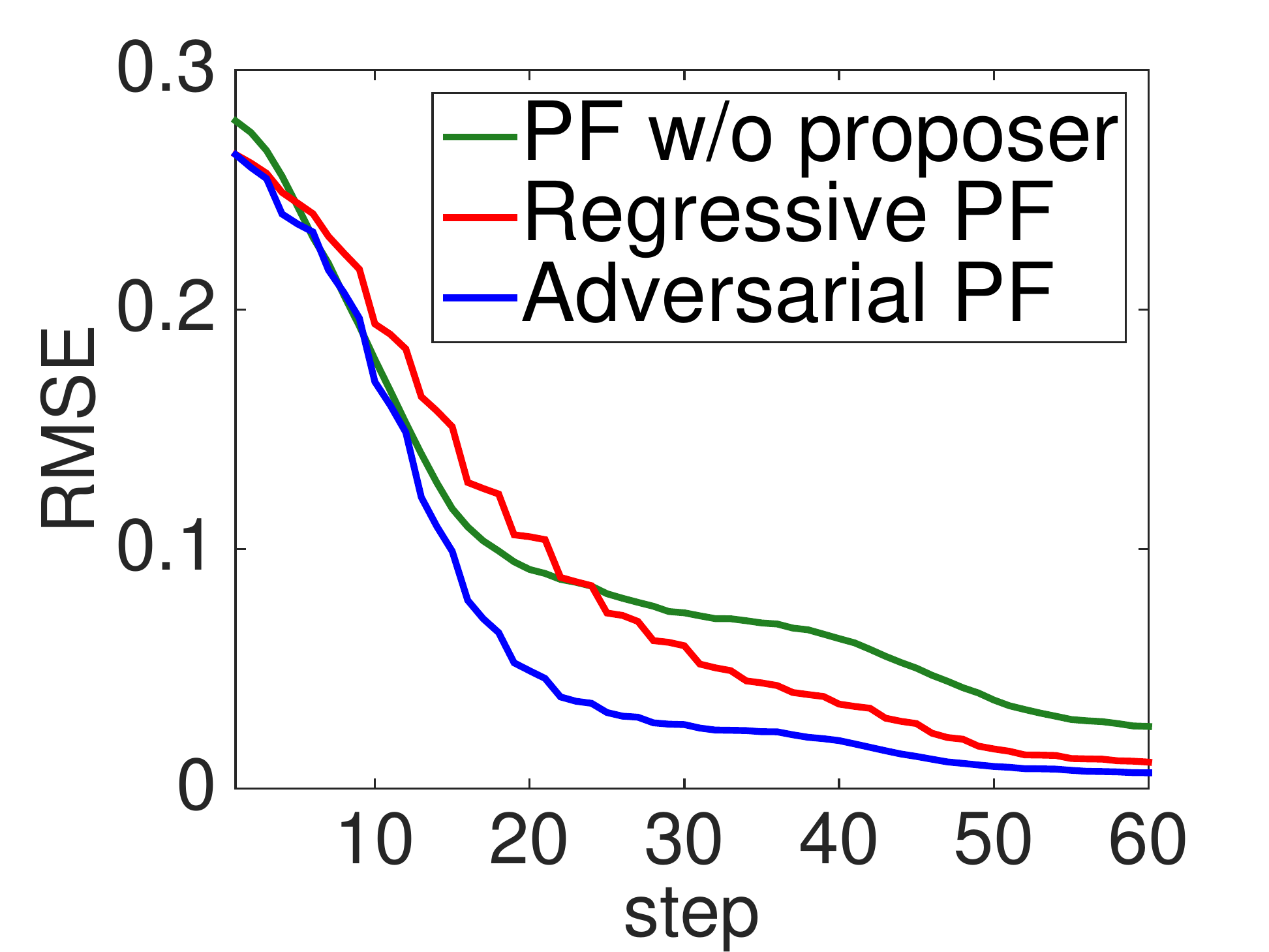}
  \label{fig:rmse_same_traj}
  }
  \vspace{-12pt}
  \caption{
  The state filtering error with respect to the number of steps which the robot has taken in the floor positioning domain
  }
  \label{fig:filter}
  \vspace{5pt}
\end{figure}

\begin{figure}[t]
  \centering
  \subfigure[Partial observation]{
  \includegraphics[width=0.47\columnwidth]{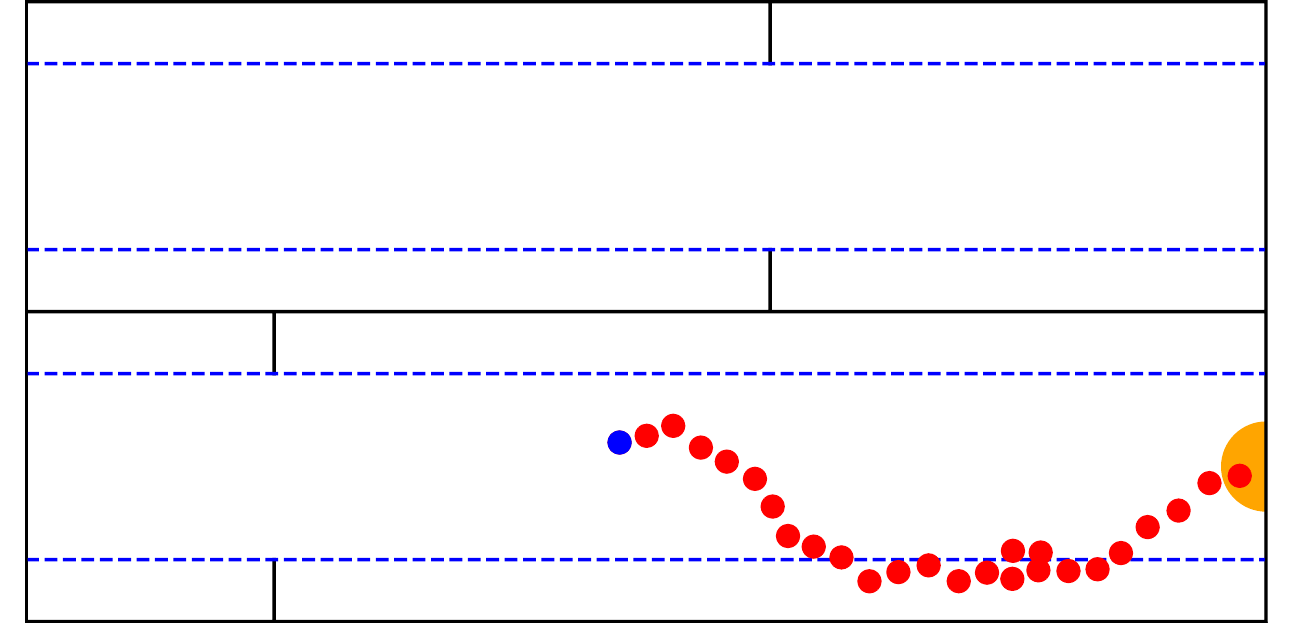}
  \label{fig:floor_pomdp}
  }
  \subfigure[Full observation]{
  \includegraphics[width=0.47\columnwidth]{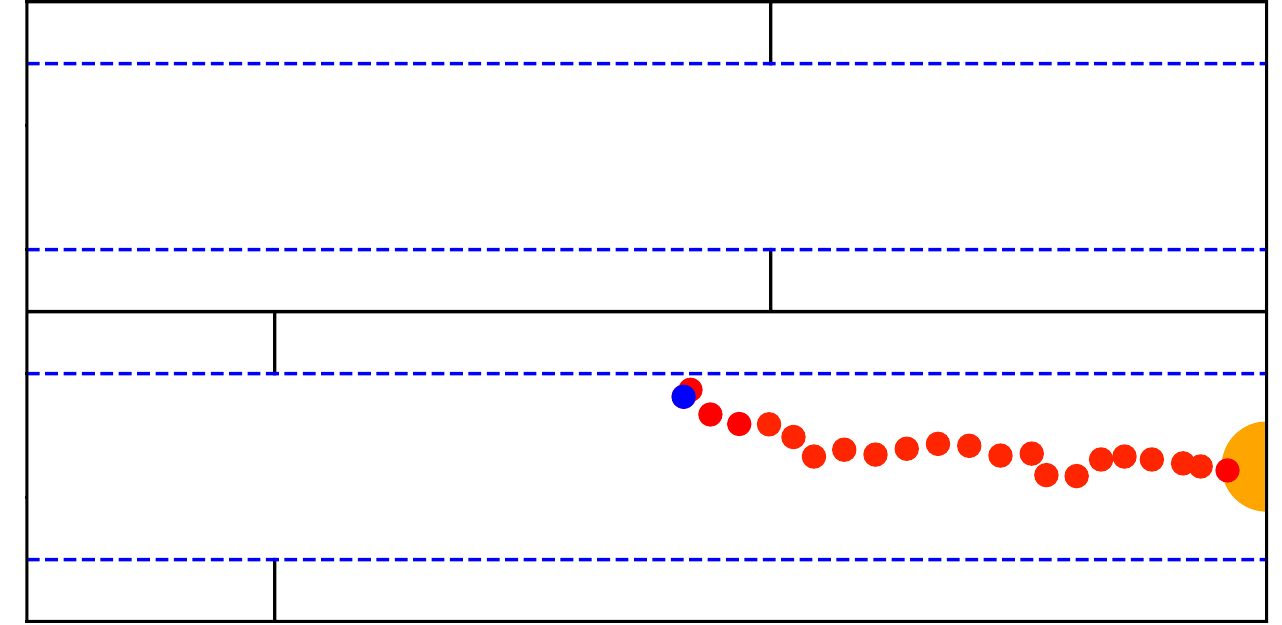}
  \label{fig:floor_true_state}
  }
  \vspace{-10pt}
  \caption{The DualSMC planner generates different polices based on the uncertainty of the perceived belief state
  }
  \vspace{5pt}
\end{figure}

\begin{figure*}[t]
  \centering
  \includegraphics[width=\textwidth]{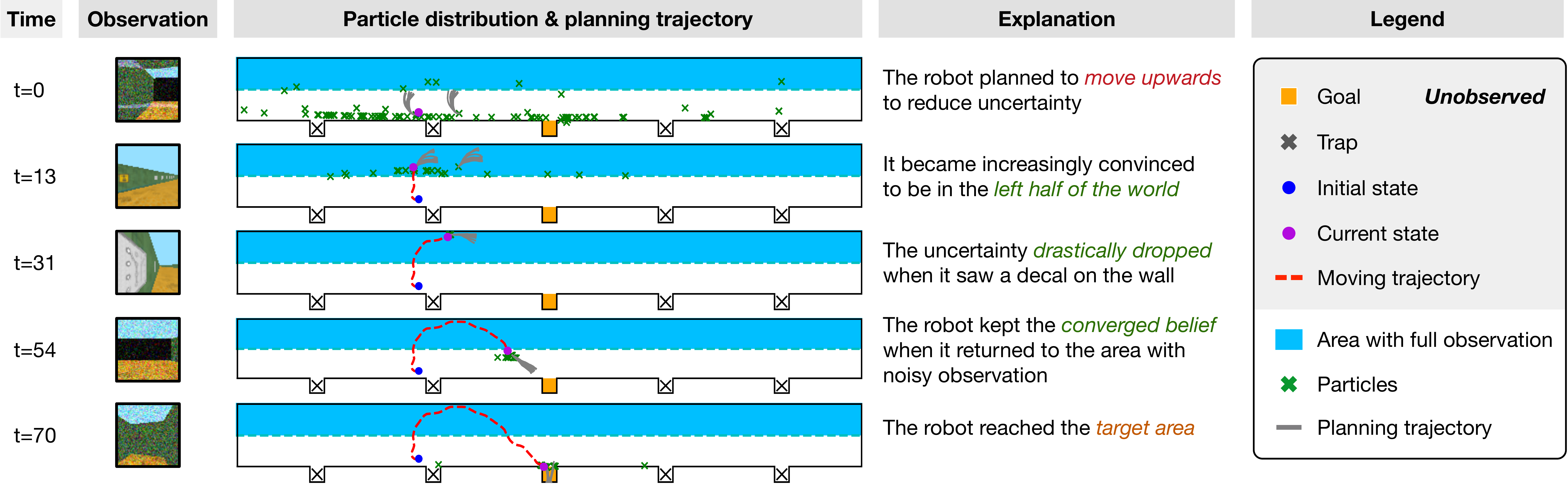}
  \vspace{-15pt}
  \caption{A demonstration trajectory from DualSMC with an \emph{adversarial filter} on the 3D light-dark navigation task
  }
  \label{fig:dmlab}
  \vspace{10pt}
\end{figure*}

\vspace{2pt}
\paragraph{Learned policies.}
The first two rows in Figure \ref{fig:floor} show the planning results by applying the SMCP algorithm \cite{piche2018probabilistic} to the top-$1$ estimated particle state.
Training the proposer with the mean squared loss is equivalent to regressing the proposed particles to the mean values of the multi-modal state distributions under partial observations. Thus, in the first example, the robot cannot make reasonable decisions due to incorrect estimation of plausible states. 
The second row in Figure \ref{fig:floor} shows the moving and planning trajectories by using an adversarial \textit{particle filter} (PF), which leads the proposed particle states closer to plausible states. The robot learns an interesting policy (along a path marked by red dots): \textit{it always goes rightwards at first, being unaware of its position until it reaches the wall, and then bouncing back at the wall}. However, this policy is suboptimal, as it does not fully consider the uncertainty of the belief state. 
In contrast, DualSMC has learned to \textit{reduce uncertainty by making short detours at first}, as shown by the last row in Figure \ref{fig:floor}. We have three findings. First, the robot learns to localize itself quickly and then approach the target area in fewer steps. 
Second, the adversarial PF works well: once the robot steps across the dashed blue line, the belief states quickly converge to the actual values, and the observation model maintains its confidence in the converged belief even when the robot moves back to the middle areas.
Third, DualSMC generates probabilistic planning trajectories of moving up/down with different advantage values.

\paragraph{Quantitative comparisons.}
From Table \ref{tab:toy}, the final DualSMC model takes $26.9$ steps to reach the target area, whilst the baseline model ``Adversarial PF + SMCP'' uses as many as $73.3$ steps on average. 
Besides, we can see that the adversarial PF significantly outperforms other differentiable state estimation approaches, such as (1) the existing DPFs that perform density estimation \cite{jonschkowski2018differentiable}, and (2) the deterministic LSTM model that was previously used as a strong baseline in \cite{karkus2018particle,jonschkowski2018differentiable}. Also note that DualSMC models with regressive proposers are even worse than one without any proposer, which suggests that an inappropriate proposer may cause a negative effect on solving continuous POMDPs.

\paragraph{Does the adversarial training improve the DPF?}
Given partial observations, an ideal filter should derive a complete distribution of possible states instead of point estimation. Figure \ref{fig:rmse_model_compare} compares the average RMSE between the true states and the filtered states by different models. The adversarial PF performs best, while the PF with the regressive proposer performs even worse than that without a proposer. A natural question arises: as the filtering error is also related to different moving trajectories of different models, can we eliminate this interference? For Figure \ref{fig:rmse_same_traj}, we train different filters without a planner. All filters follow the same expert trajectories, and the adversarial PF still achieves the best performance.

\paragraph{How does DualSMC adapt to different uncertainties?}

In a fully observable scenario, we suppress the filtering part of DualSMC and assume DualSMC plans upon a converged belief on the true state $(s_x, s_y)$. That is to say, we take the \emph{true state} to as the top-$M$ particles (line 7 in Alg \ref{alg:dualsmc}) before the planning part. The robot changes its plan from taking a detour shown in Figure \ref{fig:floor_pomdp} to walking directly toward the target area in Figure \ref{fig:floor_true_state}. It performs equally well to the standard SMCP, with a $100.0\%$ success rate and an averaged $21.3$ steps (v.s. $20.7$ steps by SMCP). We may conclude that DualSMC provides policies based on the distribution of filtered particles. We may also conclude that DualSMC trained under POMDPs generalizes well to similar tasks with less uncertainty.

\subsection{3D Light-Dark Navigation}

\begin{table}[t]
  \centering
  \small
  \begin{tabular}{lcr}
    \toprule
    Method & Success & \# Steps \\
    \midrule
    PlaNet \cite{hafner2018learning} & 30\% & 34.24 \\
    DVRL \cite{igl2018deep} & 42\% & 98.48 \\
    \midrule
    LSTM + SMCP \cite{piche2018probabilistic} & 59\% & 85.40  \\
    Adversarial PF (top-1) + SMCP & 58\%  & 56.11 \\
    Adversarial PF (top-3) + PI-SMCP & 64\% & 64.37  \\
    \midrule
    DualSMC with regressive PF ($\ell_2$) & 92\% & 66.88  \\
    DualSMC with regressive PF (density) & 98\% & 70.95  \\
    DualSMC with adversarial PF & \bf 98\% & \bf 67.49  \\
    \bottomrule
  \end{tabular}
  \vspace{-5pt}
  \caption{The average result of $100$ tests for 3D light-dark navigation
  }
  \vspace{5pt}
  \label{tab:dmlab}
\end{table}

We extend the 2D light-dark navigation domain \cite{platt2010belief} to a visually rich environment simulated by DeepMind Lab~\cite{beattie2016deepmind}. At the beginning of each episode, the robot is placed randomly and uniformly on one of the four platforms at the bottom (see Figure \ref{fig:dmlab}). The robot's goal is to navigate toward the central cave (marked in orange) while avoiding any of the four traps (marked by crosses). The maze is divided into upper and lower parts. Within the lower part, the robot travels in darkness, receives noisy visual input of a limited range (up to a fixed depth), and therefore suffers from high state uncertainty. When the robot gets to the upper part (the blue area), it has a clear view of the entire maze. We place decals as visual hints on the top walls of the maze to help the robot figure out its position. However, it has to be very close to the upper walls to see clearly what these decals are. The robot receives a positive reward of $100$ when it reaches the goal and a negative reward of $-100$ when in a trap. At each time step, the robot's observation includes a $64\times 64$ RGB image, its current velocity, and its orientation. We force it to move forward and only control its continuous orientation.

By considering the uncertainty, DualSMC methods outperform other baselines in success rate (see Table \ref{tab:dmlab}). An excessively large number of steps indicates that the robot is easy to get lost while too few steps means that it is easy to fall into a trap. From Figure \ref{fig:dmlab}, DualSMC is the only one that learned to go up and figure out its position first before going directly towards the goal.

\subsection{Modified Reacher}

We further validate our model on a continuous control task with partial observation, \ie, a modified Reacher environment from OpenAI Gym~\cite{brockman2016openai}. The original observation of Reacher is a $11$-D vector including $(\cos \theta_1, \cos \theta_2, \sin\theta_1 \sin\theta_2, g_x, g_y, \omega_1, \omega_2, r_x, r_y, r_z)$, where the first 4 dimensions are cos/sin values of the two joint angles $\theta_1$, $\theta_2$, $g_x$, $g_y$ the goal position, $\omega_1, \omega_2$ the angular velocities and $r_x, r_y, r_z$ the relative distance from the end-effector to the goal. We remove $g_x, g_y, r_x, r_y, r_z$ from the original observation and include a single scalar $r = ||(r_x, r_y, r_z)||_2 + \epsilon_r$, where $\epsilon_r \sim \mathcal{N}(0, 0.01)$ is a small noise ($r$ is usually on the scale of $0.1$). The observation is therefore a $7$-D vector. The robot has to simultaneously locate the goal and reach it.

\begin{figure}[t]
    \centering
    \includegraphics[width=0.98\columnwidth]{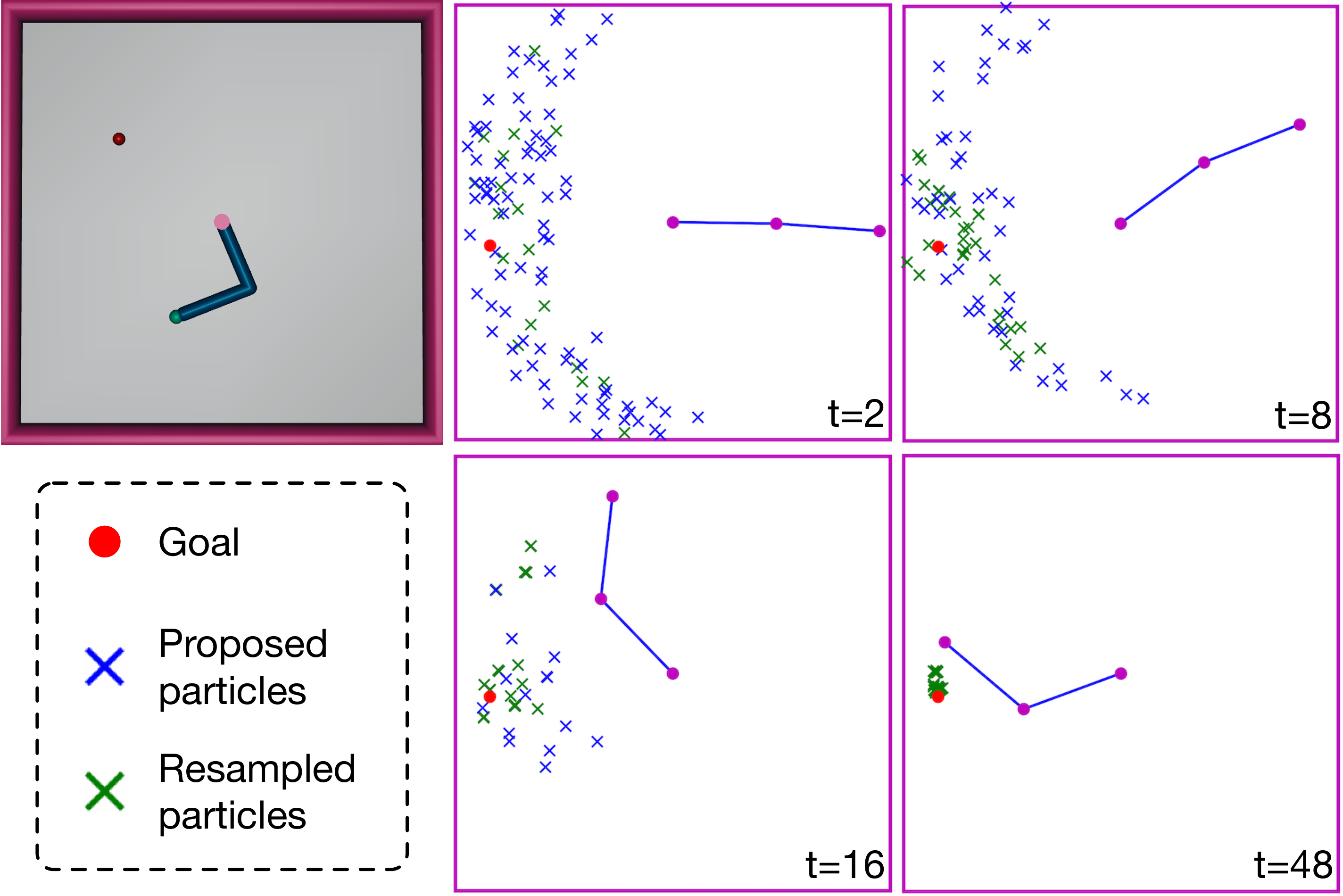}
    \vspace{-5pt}
    \caption{The modified Reacher environment and examples of the posterior belief over states given by an adversarial particle filter}
    \label{fig:reacher}
    \vspace{15pt}
\end{figure}

\begin{figure}[t]
  \centering
  \includegraphics[width=0.95\columnwidth]{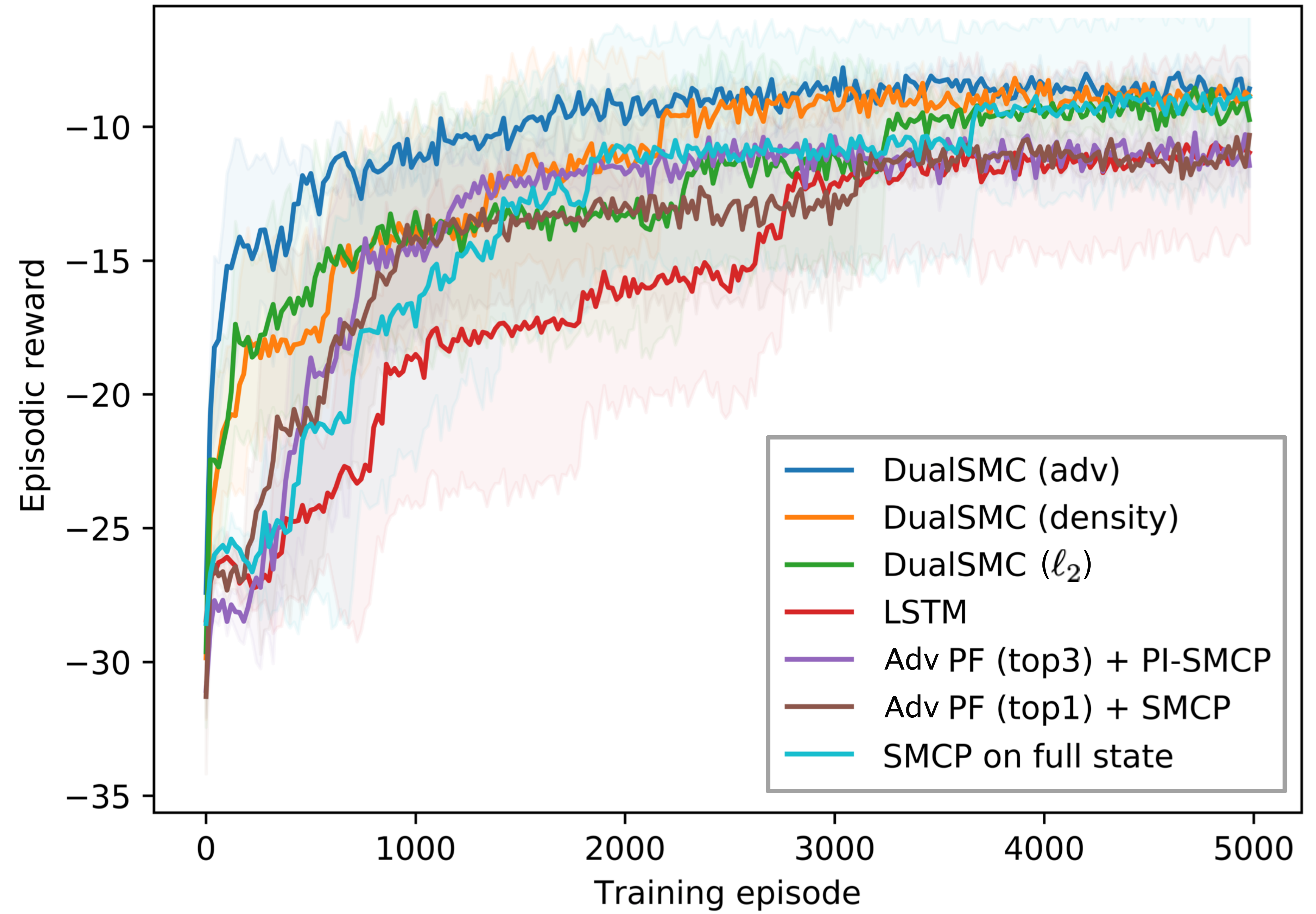}
  \vspace{-5pt}
  \caption{Training curves of DualSMC and baseline methods for the modified Reacher environment (averaged over $5$ seeds)
  }
  \label{fig:reachertrain}
  \vspace{10pt}
\end{figure}

We provide a visualization of one sample run under DualSMC with the adversarial filter in Figure \ref{fig:reacher}. As expected, initially the proposed particles roughly are in a half-cycle and as time goes on, the particles gradually concentrate around the true goal. Since the final performance of various methods is similar after long enough time of training, we provide the training curve of these methods in Figure~\ref{fig:reachertrain}, and truncate the results up to $5{,}000$ episodes since no obvious change in performance is observed from thereon. As we can see, the DualSMC methods not only achieve similar asymptotic performance as the SMCP method with full observation but also learn faster to solve the task than baseline methods.

\section{Conclusion}\label{conclusion}

In this paper, we provided an end-to-end neural network named DualSMC to solve continuous POMDPs, which has three advantages. First, it learns plausible belief states for high-dimensional POMDPs with an adversarial particle filter. For simplicity, we use the na\"ive adversarial training method from the original GANs 
\cite{goodfellow2014generative}. One may potentially improve DualSMC with modern techniques
to stabilize training and lessen mode collapse. Second, DualSMC plans future actions by considering the distributions of the learned belief states. The filter module and the planning module are jointly trained and facilitate each other. Third, DualSMC combines the richness of neural networks as well as the interpretability of classical sequential Monte Carlo methods. We empirically validated the effectiveness of DualSMC on different tasks including visual navigation and control.

\section*{Acknowledgments} 
This work is in part supported by ONR MURI N00014-16-1-2007.

\appendix
\section{Particle-Independent SMC Planning}
\label{appendix_pismcp}

As shown in Alg \ref{alg:pi_smcp}, it takes the top-$M$ particle states (for computation efficiency) and plans $N$ future trajectories \emph{independently based on each particle state}. At the end of the planning horizon $H$, it samples a trajectory from $M \times N$ planning trajectories. Although PI-SMCP is unbiased, it does not perform well in practice because it cannot generate policies based on dynamically varying state uncertainties.

\section{A Simpler Formulation of SMC Planning}
\label{appendix_dev}

\begin{figure}[H]
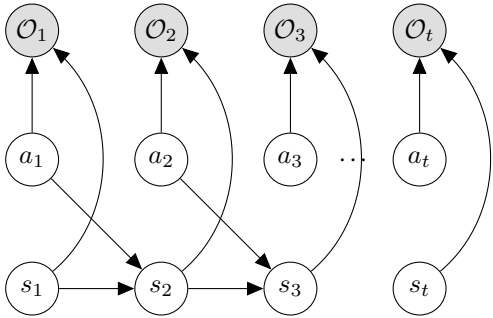

\label{smcp-proof}
  \centering
  \tikz{ %
    \node[latent] (a1) {$a_1$} ; %
    \node[latent, right=of a1] (a2) {$a_2$} ; %
    \node[latent, right=of a2] (a3) {$a_3$} ; %
    \node[latent, right=of a3] (at) {$a_t$} ; %

    \node[latent, below=of a1] (s1) {$s_1$} ; %
    \node[latent, right=of s1] (s2) {$s_2$} ; %
    \node[latent, right=of s2] (s3) {$s_3$} ; %
    \node[latent, right=of s3] (st) {$s_t$} ; %

    \node[obs, above=of a1] (o1) {$\mathcal{O}_1$} ; %
    \node[obs, right=of o1] (o2) {$\mathcal{O}_2$} ; %
    \node[obs, right=of o2] (o3) {$\mathcal{O}_3$} ; %
    \node[obs, right=of o3] (ot) {$\mathcal{O}_t$} ; %

    \edge {a1} {o1} ; %
    \edge {a2} {o2} ; %
    \edge {a3} {o3} ; %
    \edge {at} {ot} ; %
    \edge {a1} {s2} ; %
    \edge {a2} {s3} ; %
    \edge {s1} {s2} ; %
    \edge {s2} {s3} ; %

    \path (s1) edge [bend right=50,->]  (o1) ; %
    \path (s2) edge [bend right=50,->]  (o2) ; %
    \path (s3) edge [bend right=50,->]  (o3) ; %
    \path (st) edge [bend right=50,->]  (ot) ; %
    \path (a3) -- node[auto=false]{\ldots} (at);
  }
  \vspace{-5pt}
  \caption{$\mathcal{O}_t$ is the observed \textit{optimality} variable with probability $p(\mathcal{O}_t|s_t, a_t) = \exp(r(s_t, a_t))$, where $r(s, a)$ is the reward function}
  \label{fig:graphical_model}
  \vspace{10pt}
\end{figure}

At time $t$, we set $Q_\omega(\hat{s}_{i-1}^{(m)(n)}, a_{i-1}^{(n)})$ and $r_{i-1}^{(m)(n)}$ in Eq. \eqref{eq:advantage} to $0$. We emphasize that our formulation is much simpler than the original SMCP \cite{piche2018probabilistic}. Because it only requires a learned $Q$ function and more importantly, it prevents us from estimating the expectation of the value function $V$. 
To prove this, we depict the Hidden Markov Model of our planning algorithm for ease of notation. Figure \ref{fig:graphical_model} is borrowed from \cite{piche2018probabilistic}. $\mathcal{O}_t$ is a convenience binary variable here for the sake of modeling, denoting the ``optimality'' (optimal policy) of a pair $(s_t, a_t)$ at time $t$ \cite{levine2018reinforcement}.
Then we present the derivation of line 8 Alg \ref{alg:dualsmc-p} as follows. Comparing with  \cite{piche2018probabilistic}, our update of the planning particle weights depends only on the learned $Q$ and $\pi_\rho$.

\begin{table*}[t]
\vspace{10pt}
    \centering
    \small
    \begin{tabular}{c|ll|llll}
    \toprule
    Module & Layers (floor positioning) & \# Channels & Layers (3D dark-light) & \# Channels & Filter & Stride \\
    \midrule
    $Z_\theta$  
    & $3$-layer MLP & $256\times2, 16$ 
    & Conv2d$\times2$, MaxPool($2$) & $16, 32$ & $(3,3)$ & $2$
    \\
    & & & Conv2d, Dropout($0.2$) (*) & $64$ & $(3,3)$ & $2$ \\
    & & & Fully connected & $64$ \\
    & $2$-layer LSTM & $128\times2$ & $2$-layer LSTM & $64\times2$ \\
    & & & Concat: state, orientation & $67$ \\
    & $3$-layer MLP & $256\times2, 1$ & $4$-layer MLP & $128\times3, 1$ \\
    $P_\phi$ 
    & $3$-layer MLP & $256\times2, 64$ & Same as $Z_\theta$ up to (*) & $64$ & $(3,3)$ & $2$ \\
    & & & Fully connected & $64$ \\
    & Concat: $z (64) \sim \mathcal{N}(0,1)$ & $128$ 
    & Concat: $z (64) \sim \mathcal{N}(0,1)$, orientation & 129 \\
    & $4$-layer MLP & $256\times3, 2$  
    & $3$-layer MLP & $128\times3, 2$ \\
    $T_\psi$  
    & $4$-layer MLP & $256\times3, 4$ 
    & Action noise $\sim \mathcal{N}(0,1)$ & $1$ \\
    & & & $3$-layer MLP: encode action noise to $e$ & $128\times2, 1$ \\
    & & & Concat: (state, action + $e$) & $6$ \\
    & & & $3$-layer MLP, then add to state & $128\times3, 5$ \\
    $Q_\omega$ 
    & $3$-layer MLP & $256\times2, 1$ 
    & $3$-layer MLP & $128\times2, 1$ \\
    $\pi_\rho$ 
    & $3$-layer MLP & $256\times2, 4$
    & $3$-layer MLP & $128\times2, 1$ \\
    \bottomrule
    \end{tabular}
    \vspace{-5pt}
    \caption{Network details of each module in DualSMC}
    \vspace{5pt}
    \label{tab:model}
\end{table*}

\section{Network Details}
\label{appendix_model}

Table \ref{tab:model} shows the network details for the floor positioning domain and the 3D dark-light domain.

\begin{algorithm}[h]
\caption{Particle-Independent SMC Planning}
\label{alg:pi_smcp}
\textbf{Input:} $\{\tilde{s}_t^{(m)}, \tilde{w}_t^{(m)}\}_{m=1}^M$ \\
\textbf{Output}: $a_t$
\begin{algorithmic}[1]
\STATE $\{\hat{s}_t^{(m \times n)} = \tilde{s}_t^{(m)}, {\hat{w}_t^{(m \times n)} = \tilde{w}_t^{(m)}}\}_{m=1,n=1}^{M,N}$
\STATE $\{\hat{w}_t^{(n)}\}_{n=1}^{M N} = \text{Normalize}(\{\hat{w}_t^{(n)}\}_{n=1}^{M N})$
\vspace{2pt}
\FOR{$i = t:t+H$}
    \STATE // Predict actions based on individual particle states
    \STATE $\{a_i^{(n)} \sim \pi_\rho (\hat{s}_i^{(n)})\}_{n=1}^{M N}$
    \STATE $\{\hat{s}_{i+1}^{(n)}, r_i^{(n)} \sim T_\psi(\hat{s}_i^{(n)}, a_i^{(n)})\}_{n=1}^{M N}$
    \STATE $\{\hat{w}_{i+1}^{(n)} \propto \hat{w}_i^{(n)} \cdot \exp(A(\hat{s}_i^{(n)}, a_i^{(n)}, \hat{s}_{i+1}^{(n)}))\}_{n=1}^{M N}$
    \STATE $\{x_i^{(n)} = (\hat{s}_{i+1}^{(n)}, a_i^{(n)}, \hat{s}_i^{(n)})\}_{n=1}^{M N}$
    \vspace{2pt}
    \IF{resample}
    \STATE $\{x_{t:i}^{(n)}\}_{n=1}^{M N} \sim \text{Multinomial}(\{x_{t:i}^{(n)}\}_{n=1}^{M N}), \text{\wrt} \{\hat{w}_{i+1}^{(n)}\}_n$
    \vspace{-7pt}
    \STATE $\{\hat{w}_{i+1}^{(n)}=1\}_{n=1}^{M N}$
    \vspace{2pt}
    \ENDIF
\ENDFOR
\STATE $a_t =$ first action of $x_{t:t+H}^{(n)}$, $n \sim \text{Uniform}(1, \dots, MN)$
\end{algorithmic}
\end{algorithm}

\vspace{-8pt}

\begin{equation}
\small
\begin{split}
w_t &= \frac{p(x_{1:t} | \mathcal{O}_{1:T})}{q(x_{1:t})} \\
&= \frac{p(x_{1:t-1}|\mathcal{O}_{1:T})}{q(x_{1:t-1})}\frac{p(x_t|x_{1:t-1}, \mathcal{O}_{1:T})}{q(x_t|x_{1:t-1})}\\
&= w_{t-1} \frac{p(x_t|x_{1:t-1}, \mathcal{O}_{1:T})}{q(x_t|x_{1:t-1})}\\
&= \frac{w_{t-1}}{q(x_t|x_{1:t-1})} \frac{p(x_{1:t}|\mathcal{O}_{1:T})}{p(x_{1:t-1} | \mathcal{O}_{1:T})}\\
&= \frac{w_{t-1}}{q(x_t|x_{1:t-1})} \frac{p(\mathcal{O}_{1:T} | x_{1:t})p(x_{1:t})}{p(\mathcal{O}_{1:T} | x_{1:t-1})p(x_{1:t-1})}\\
&= \frac{w_{t-1}}{q(x_t|x_{1:t-1})} \frac{p(\mathcal{O}_{1:t-1} | x_{1:t-1})p(x_{1:t})p(\mathcal{O}_{t:T}|x_t)}{p(\mathcal{O}_{1:t-2} | x_{1:t-2})p(x_{1:t-1})p(\mathcal{O}_{t-1:T} |x_{t-1})}\\
&= \frac{w_{t-1}}{q(x_t|x_{1:t-1})} p(x_t | x_{t-1}) p(\mathcal{O}_{t-1} | x_{t-1}) \\
&\quad \exp\big(Q(s_t, a_t) - Q(s_{t-1}, a_{t-1})\big)\\
&= w_{t-1}\frac{p(x_t|x_{t-1})}{q(x_t|x_{t-1})} 
\exp\big(Q(s_t, a_t) - Q(s_{t-1}, a_{t-1}) + r_{t-1}\big)\\
&= w_{t-1}\frac{p_{env}(s_t|s_{t-1}, a_{t-1})}{p_{model}(s_t|s_{t-1}, a_{t-1})} \\
&\quad \exp\big(Q(s_t, a_t) - Q(s_{t-1}, a_{t-1}) + r_{t-1} - \log \pi_\rho (a_t | s_t)\big).\\
\end{split}
\end{equation}

\bibliographystyle{named}
\bibliography{ijcai20}

\end{document}